\documentclass[letterpaper, 10 pt, conference]{ieeeconf}  % Comment this line out
\IEEEoverridecommandlockouts                              % This command is only
\usepackage{cite}
\usepackage{amsmath,amssymb,amsfonts}
\usepackage{algorithmic}
\usepackage{graphicx}
\usepackage{textcomp}
\usepackage{xcolor}
\usepackage{kotex}
\usepackage{amsmath}
\usepackage{multirow}
\usepackage{float}
\usepackage{bigstrut}
\usepackage{url}
\usepackage[english]{babel}

\title{\LARGE \bf
Fisheye-Lens-Camera-based Autonomous Valet Parking System
}

\author{Young Gon Jo, Seok Hyeon Hong, Student Member, IEEE, Sung Soo Hwang, and Jeong Mok Ha, Member, IEEE% <-this % stops a space
\thanks{The work and experiment were supported by VADAS Co., Ltd. English
language editing of this paper was supported by Editage (www.editage.co.kr).
This work was supported by the National Program for Excellence in Software
at Handong Global University (2017-0-00130) fundeded by the Ministry of
Science and ICT. (Young Gon Jo and Seok Hyeon Hong contributed equally
to this work.) (Corresponding author: Sung Soo Hwang)}% <-this % stops a space
\thanks{Young Gon Jo graduated from Handong University with a B.S. degree in
computer and electrical engineering. He is now working at VADAS Co., Ltd.
(e-mail: 21500690@handong.edu).}%
\thanks{Seok Hyeon Hong is with the School of Computer Science and Electrical
Engineering, Handong University, Pohang 37554, South Korea (e-mail:
21500798@handong.edu).}%
\thanks{Sung Soo Hwang graduated from the Korea Advanced Institute of Science
and Technology with a doctorate in electrical and electronic engineering. He is
now an associate professor at Handong University in the School of Computer
Science and Electrical Engineering (e-mail: sshwang@handong.edu).}%
\thanks{Jeong Mok Ha graduated from Pohang University of Science and Technology
in Electrical Engineering. He is now the head of the algorithm team at
VADAS Co., Ltd. (e-mail: jmha@vadas.co.kr).}%
}

\begin{document}

\maketitle
\thispagestyle{empty}
\pagestyle{empty}

%%%%%%%%%%%%%%%%%%%%%%%%%%%%%%%%%%%%%%%%%%%%%%%%%%%%%%%%%%%%%%%%%%%%%%%%%%%%%%%%
\begin{abstract}

This paper proposes an efficient autonomous valet
parking system utilizing only cameras which are the most widely
used sensor. To capture more information instantaneously and
respond rapidly to changes in the surrounding environment,
fisheye cameras—which have a wider angle of view compared
to pinhole cameras—are used. Accordingly, visual simultaneous
localization and mapping (SLAM) is used to identify the layout
of the parking lot and track the location of the vehicle. In
addition, the input image frames are converted into around-viewmonitor
(AVM) images to resolve the distortion of fisheye lens
because the algorithm to detect edges are supposed to be applied
to images taken with pinhole cameras. The proposed system
adopts a look-up table for real-time operation by minimizing
the computational complexity encountered when processing AVM
images. The detection rate of each process and the success rate
of autonomous parking were measured to evaluate performance.
The experimental results confirm that autonomous parking can
be achieved using only visual sensors.
\end{abstract}


%%%%%%%%%%%%%%%%%%%%%%%%%%%%%%%%%%%%%%%%%%%%%%%%%%%%%%%%%%%%%%%%%%%%%%%%%%%%%%%%
\section{INTRODUCTION}

AS self-driving technology continues to develop with
the aim of absolutely unaided driving, advanced driverassistance
systems (ADAS) are likewise developing with the
goal of complete automation, which is the fifth level of
automation defined by the Society of Automotive Engineers(SAE).

Typical ADAS technology includes an autonomous parking
system that aids drivers in precise and facile parking. Developed
and presented jointly by German automotive multinationals
Daimler AG and Robert Bosch, the autonomous
valet parking (AVP) system allows the vehicle to be parked
and recalled through a dedicated smartphone application. In
the case of Hyundai’s and Kia’s automatic parking support systems, the speed and transmission are operated by the vehicle itself via ultrasonic sensors attached to the vehicle.

However, there are still many problems with existing autonomous
parking systems. For instance, AVP systems (Daimler
AG and Bosch) provide parking-space information to
self-parking vehicles by installing sensors in the parking lot
itself; these sensors are expensive to install and difficult to
commercialize because the core of the system is in the parking
lot itself. In addition, in AVP, the driver must first move the car
directly into the parking lot and near the preferred parking bay,
providing only simple functions such as basic reversing. On
the other hand, self-parking systems using ultrasonic sensors
are limited in that parking is impossible when the lot is empty;
these systems cannot obtain information about the vehicle’s
vicinity when there is no parked vehicle for reference.

Therefore, in this study, we propose a automated AVP
system utilizing efficient path-generation and driving strategies
in parking lots where lane and signal information are unclear.
Essentially, the system improves general functionality at a
lower cost. For instance, should the driver disembark at any
location and run the proposed system, the car will navigate
into the parking lot independently; if an empty bay is located,
the car will park unaided. The novel system uses neither light
detection and ranging (LiDAR) sensors that otherwise hinder
commercialization due to expensive unit prices [1], nor GPS
sensors that cannot obtain information inside the parking lot
except location information [2], nor ultrasonic sensors that
only measure the distance from nearby objects, and are hence
less accurate. Instead, similar to the method proposed in [3],
this system performs autonomous parking using a fisheye
camera and a sensor that accurately identifies information in
the indoor parking lot while having an affordable unit price.
\section{PROPOSED DESIGN}

\subsection{Overall System Architecture}

Four camera sensors are attached in four directions (front,
rear, left, and right) as in Fig. 1. Visual simultaneous localization
and mapping (SLAM) is run using footage captured
by the front camera to map the layout of the parking lot and
simultaneously pin-point the current location of the vehicle.
The front camera is also used to detect signs such as direction
indicator arrows. The other sensors are used to detect empty
parking bays and to accordingly execute the parking algorithm.

Monocular pinhole lenses have an angle of view narrower
than 50°, making it difficult to obtain information about the
vicinity, such as parking lines and obstacles on the pavement.
In this regard, the proposed system uses fisheye lenses with an angle of view of 180° to capture more information through the cameras.

\begin{figure}[htp] 
    \centering 
    \includegraphics[width=200pt]{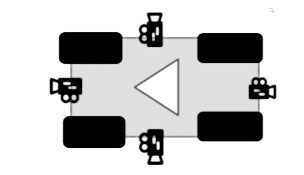} 
    \caption{Sensor layout used in the system} 
    \label{fig:fig1}
\end{figure}

While fisheye lenses have the advantage of having a wide
angle of view, they have the disadvantage of distortion in
images as illustrated in Fig. 2 In other words, visual SLAMs
based on ordinary lenses, such as ORB SLAM [4], cannot be
used. Consequently, the process of correcting the distortion
of fisheye-lens images is necessary to utilize the information
captured. To this end, location estimation and mapping are
performed using CubemapSLAM [5] to resolve distortions.
CubemapSLAM is based on a previous study, which asserts
that feature extraction and pattern recognition are possible in
images captured via a fisheye lens [6]. Moreover, Cubemap-
SLAM uses ORB feature, which is more efficient compared
to SIFT and SURF [7]. As depicted in Fig. 3, the image is
projected onto the cube to correct the distortion, and features
are extracted from the unfolded cube image.

\begin{figure}[htp] 
    \centering 
    \includegraphics[width=120pt]{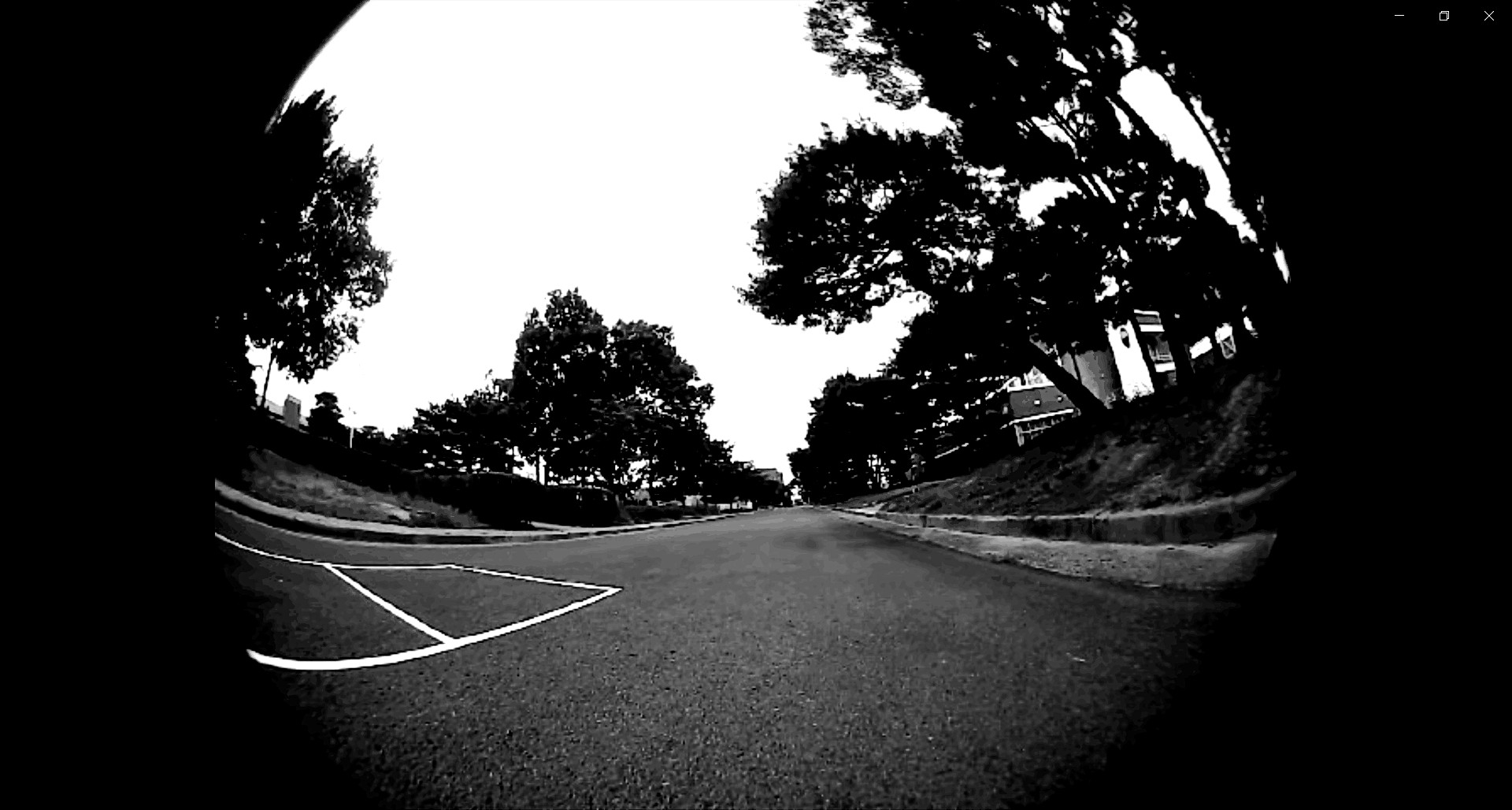} 
    \caption{Video captured by the Fisheye-lens camera} 
    \label{fig:fig2}
\end{figure}

\begin{figure}[htp] 
    \centering 
    \includegraphics[width=120pt]{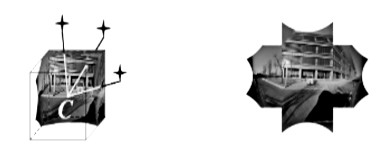} 
    \caption{Converted CubemapSLAM image} 
    \label{fig:fig3}
\end{figure}

Fig. 4 is a brief flow chart of the system, which is largely
divided into four parts: initial self-driving before loop closing,
keyframe-based self-driving after loop closing, autonomous
parking, and return to the spot where the driver calls from.

The first step (self-driving before loop closing) involves
analyzing the images through the camera sensors. When the driving image frame is input into the system, the system starts the initial autonomous driving. Straight driving is the default
setting for the initial driving, and the vehicle is steered by
identifying road markings in the parking lot while driving. In
this phase, the vehicle also constructs the SLAM map by going
around the parking lot.

\begin{figure}[htp] 
    \centering 
    \includegraphics[width=240pt]{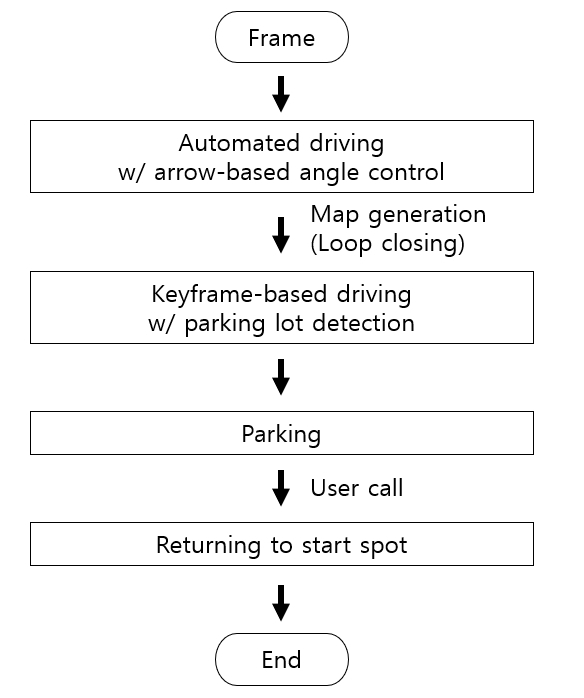} 
    \caption{Flow diagram of autonomous valet parking system} 
    \label{fig:fig4}
\end{figure}

If the vehicle arrives at the start position, the vehicle does
not repeat the identification phase; instead, it drives along
the keyframe in the SLAM map. Each keyframe has its
own unique identification number in the order in which it
is generated, and sorting based on this number allows it to
be accessed sequentially. Loop closing is one of the SLAM
processes that corrects cumulative errors by evaluating that a
loop is created upon revisiting an area of the parking lot; loop
closing facilitates precise three-dimensional (3D) mapping
with minimal space error. This implies that the route traversed
by the vehicle also becomes more accurately defined. Thus, it
is possible to drive unaided in the order in which the keyframes
were generated.

If a vacant parking bay is found, the vehicle is driven to the
appropriate location and the parking algorithm is executed.
The location that the vehicle shall be moved to is indicated
in the SLAM map; the real-time location of the vehicle
is consistently tracked by checking whether it is moving
appropriately.

\subsection{Detailed Design}
1) Autonomous Driving before Loop Closing

Owing to lack of information on the layout of the parking
lot prior to loop closing, the driving direction and steering of
the vehicle should be determined by information found in the
parking lot. We developed the system under the assumption
that the parking lot has markings (arrows) on the floor that
indicate the direction in which the vehicle should proceed.

The head and tail of the arrow are detected to identify the
direction that the arrow indicates[8]. The arrows used in this
study were left and right arrows; the gradient of the two edges
of the arrow head is detected and the center of each edge is
compared to increase the accuracy of the arrow identification.
For example, the left arrow is identified if the center of the
edge with a positive gradient is above the center of the edge
with a negative gradient as shown in Fig. 5, and vice versa. To
prevent erroneous identification cases due to noise, the rotation
control command was applied when more than 10 frames were
continuously identified in the same direction. Moreover, arrow
identification is not performed while the vehicle is rotating to
allow the vehicle to smoothly rotate by 90°.

\begin{figure}[htp] 
    \centering 
    \includegraphics[width=240pt]{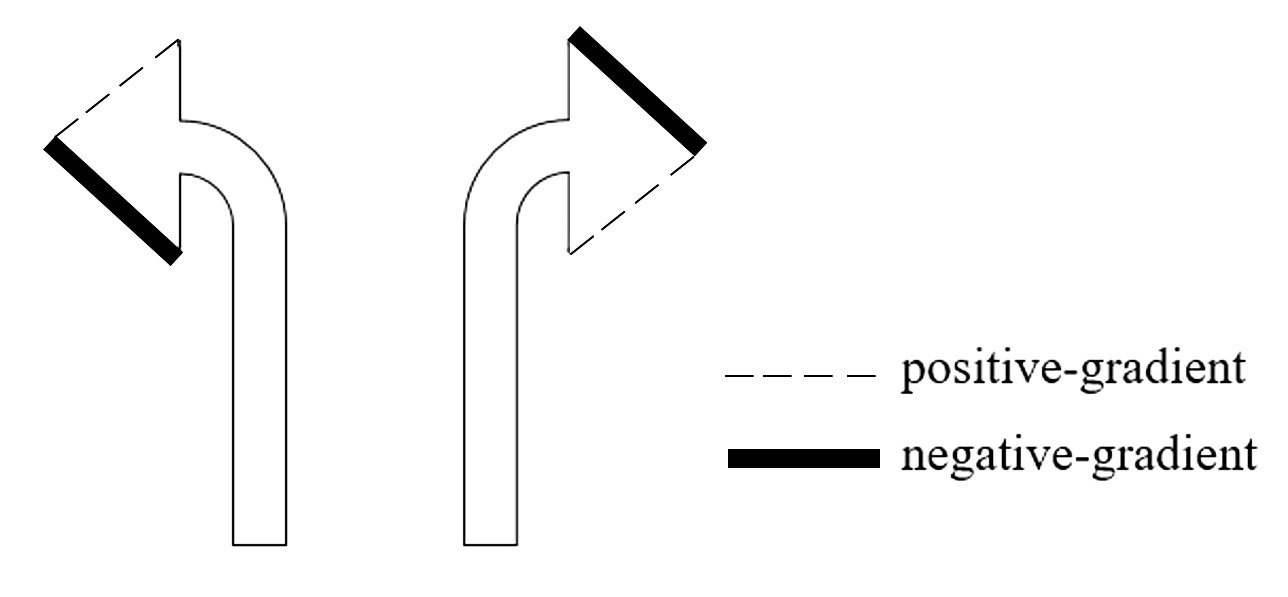} 
    \caption{Classification of direction of arrows} 
    \label{fig:fig5}
\end{figure}

2) Autonomous Driving after Loop Closing

After loop closing occurs, the keyframes are sorted in
the order in which they were generated, and the keyframes
currently in front of the vehicle are followed as the vehicle
retraces its route. The vehicle’s coordinate system is used
to locate the target keyframe; this keyframe determines the
direction followed by the vehicle as well as steering angles.

To ensure that the vehicle reaches the target keyframe, it is
necessary to check the distance between the vehicle and the
target keyframe. The arrival criteria are set as the average of
the distances between adjacent keyframes. Nonetheless, if the
target keyframe is located behind the vehicle, it is assumed
that the vehicle has already passed that keyframe.

Exception handling is required when the vehicle cannot autonomously
reach the target keyframe position. The values of
y-axis representing height on the 3D map have no significant
effect on determining direction. Therefore, the position of all
keyframes is projected onto the xz-plane. The equation of a
circle passing through all three adjacent coordinates (x1, z1),
(x2, z2), and (x3, z3) of the projected position is obtained
using equation 1 and 2.

The minimum value of the radius obtained using equation 1
is set as the turning radius of the vehicle, as well as the radius
of the unreachable area as shown in Fig. 6.

If the subsequent keyframe is located in the unreachable
area, the vehicle applies corrective maneuvers. For example,
if the keyframe is located on the left side, the vehicle reverses
to the right until the keyframe is out of the unreachable area,
and vice versa.

\begin{figure}[htp] 
    \centering 
    \includegraphics[width=240pt]{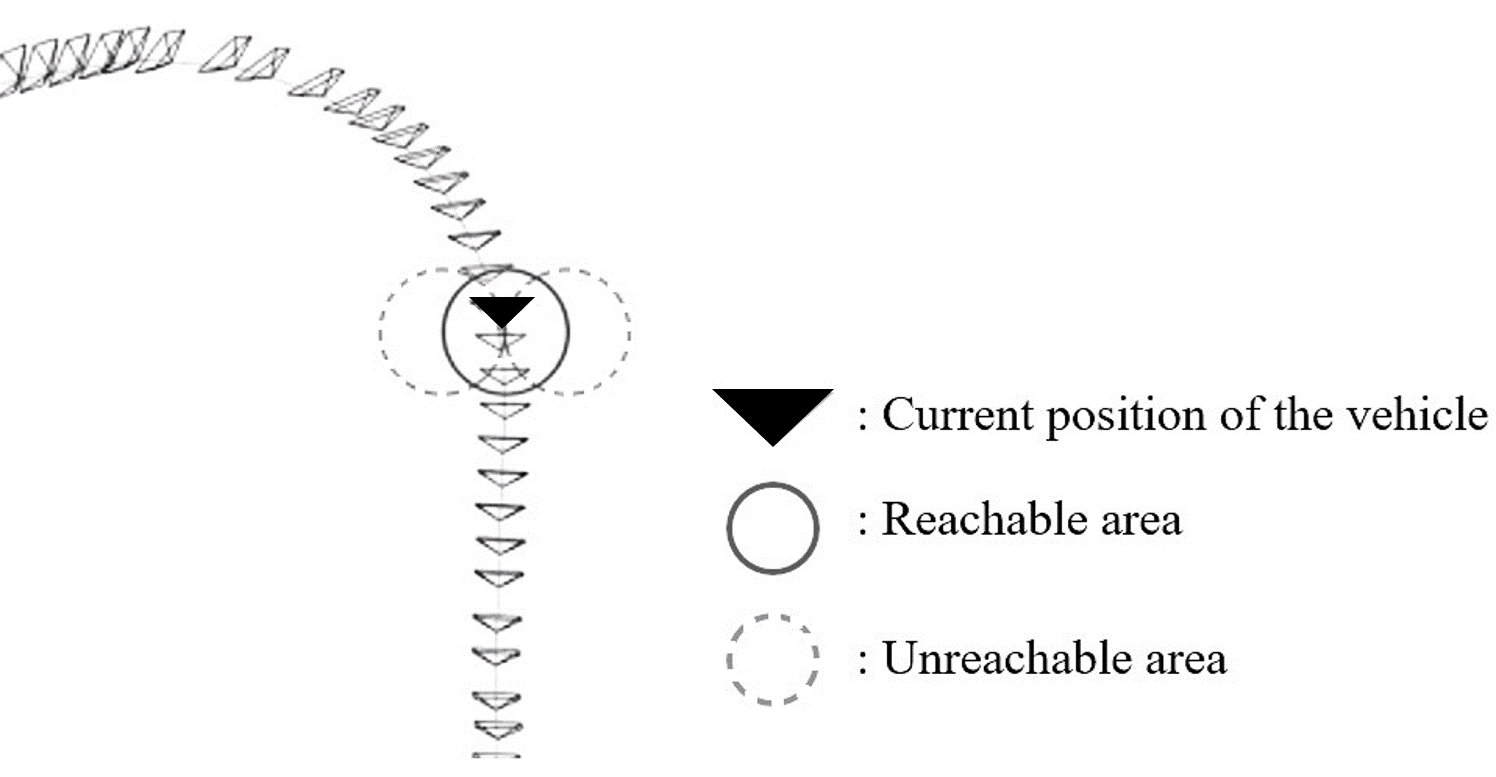} 
    \caption{SLAM map that shows the result of the reachability} 
    \label{fig:fig6}
\end{figure}

3) Empty Parking Space Identification

In general, parking lines are detected to identify empty
parking spaces in the video, and parking spaces are identified
using the angle of parking lines, the separation between
lines, location of lines, and so on. Such parameters generally
limit accurate detection of parking spaces to particular
environments. The proposed system reinforces existing linedetection
methods and proposes an adaptive parking-space
detection method using template matching techniques [12].
These methods can detect parking spaces of various shapes
without numerical constraints and reduce cases of erroneous
detection.

\begin{figure}[htp] 
    \centering 
    \includegraphics[width=240pt]{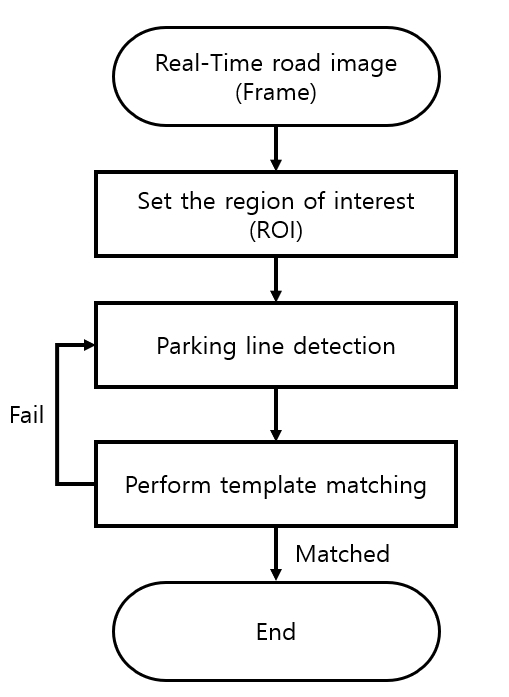} 
    \caption{Flowchart of detecting empty parking spaces} 
    \label{fig:fig7}
\end{figure}

Fig. 7 shows the sequence of empty-parking-space detection
algorithms. When loop closing occurs while driving, the
camera receives images from the front, rear, left, and right
of the vehicle as system input. Given that parking lot images
typically contain a large number of lines, detecting lines for all
pixels in the image reduces the accuracy of detection and also
puts a strain on the system in terms of computational speed.
To solve this problem, the proposed system has a preliminary
process that uses only a subset of the images by designating a
region of interest (ROI). Thereafter, the relevant parking line
and the corresponding parking bay are detected in the area of
interest. The detected line segment is used as the target image
in the template-matching technique. A software that finds
similarities between the target image and the template is then
utilized to identify the parking space. If the two images are
deemed similar, the corresponding area is positively identified
as a parking space and the parking algorithm is executed.
However, in case of failure, the driving continues and the other
parking spaces are explored.

To detect parking spaces, parking lines demarcating parking
spaces must first be detected. OpenCV supports several linesegment
detection algorithms. Nevertheless, conventional linesegment
detection algorithms such as Houghline [9, 10] are
adversely affected by environmental changes—each change
in image environment necessitates modification of the value
provided as a function factor. In addition, because such an
algorithm detects one long line with several short/broken lines
as well as non-parking lines, it is detrimental to use only
conventional line-segment detection methods. Therefore, for
more accurate and adaptive line detection in this system, the
line-filtering technique [11] was used to detect the edges;
the detected edges were in turn used to detect corresponding
parking lines.

Line filtering software makes parking lines brighter than
their surroundings. Upon applying the horizontal and vertical
line filters (equation 3 and 4, respectively) to one point in
the image, the calculated values are used to obtain a specific
point brighter than the surrounding area. L(x; y) in the linefilter
equations denotes the intensity (brightness) value at the
coordinates (x, y).

\begin{figure}[htp] 
    \centering 
    \includegraphics[width=240pt]{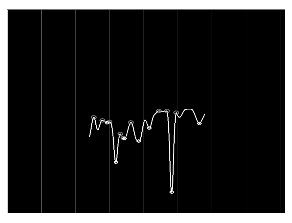} 
    \caption{Graph of the pixel value multiplied by the weight} 
    \label{fig:fig8}
\end{figure}

The graph in Fig. 8 is the result of applying a horizontal
line filter for any y-value. Given the large difference between
adjacent pixels and brightness values, the boundaries of the
parking line display a large height difference on the graph.
The median values of peak and valley above the threshold are
used as definite points as shown in Fig. 9. In this manner,
the line filter is applied to all pixels in the ROI to obtain the
specific points, and these points are used as edges to detect the
parking lines. A parking space typically consists of two split
lines and one base line, but sometimes only two split lines. In
this system, additional base lines in the parking space were
detected using the identified split lines. The split lines were
also used to detect parking spaces regardless of the baseline.

\begin{figure}[htp] 
    \centering 
    \includegraphics[width=240pt]{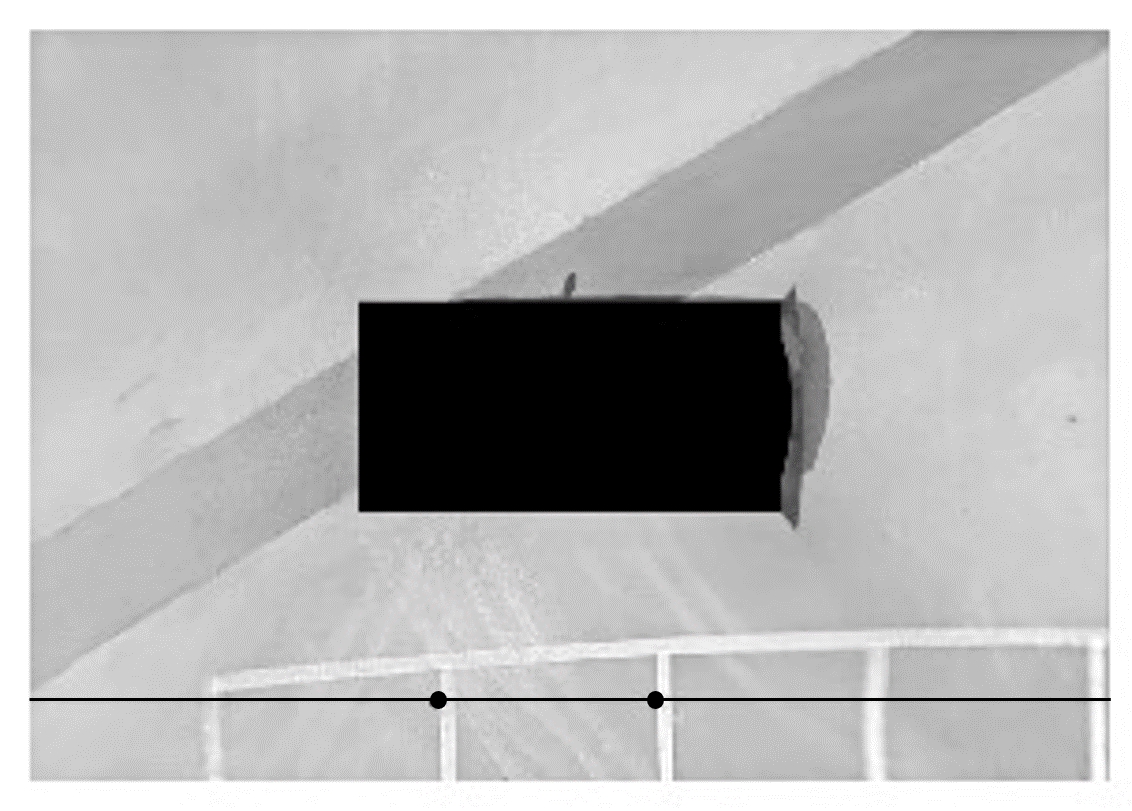} 
    \caption{Image showing peak and valley center values} 
    \label{fig:fig9}
\end{figure}

\begin{figure}[htp] 
    \centering 
    \includegraphics[width=240pt]{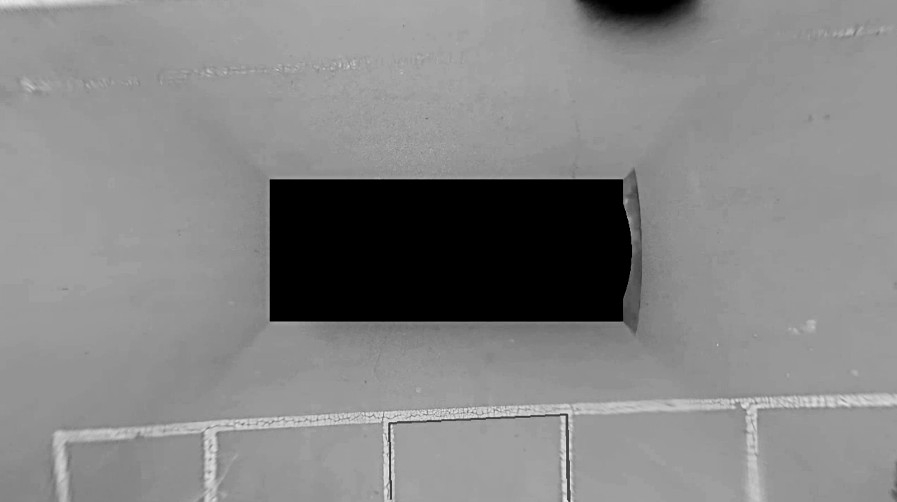} 
    \caption{Result of line detection} 
    \label{fig:fig10}
\end{figure}

The template-matching technique was used to determine
whether the area of the detected line composes the parking
space. Template matching is a technology that determines if
there is an area that matches the template image in the target
image. In order to perform template matching, a parkingspace
database must first be established. In this system,
parking-space images of various shapes were compiled into
a template database and easily compared to target images.
After template matching, the vehicle positioning and parking
steps were carried out. Fig. 13 shows the result of template
matching.

\begin{figure}[htp] 
    \centering 
    \includegraphics[width=240pt]{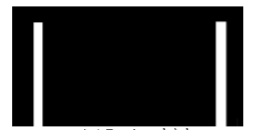} 
    \caption{Template image} 
    \label{fig:fig11}
\end{figure}

\begin{figure}[htp] 
    \centering 
    \includegraphics[width=240pt]{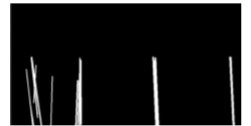} 
    \caption{Target image} 
    \label{fig:fig12}
\end{figure}

\begin{figure}[htp] 
    \centering 
    \includegraphics[width=240pt]{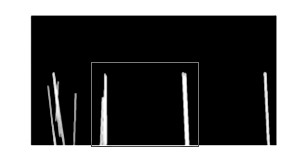} 
    \caption{Result of template matching} 
    \label{fig:fig13}
\end{figure}

4) Look-Up Table (LUT)

Because the camera attached to the vehicle was a fisheye
camera, it was impossible to apply algorithms to detect edges
and lines in images of parking lots due to severe image
distortion. Therefore, to identify the parking space, we corrected
the distortion of the image acquired from the four-way
camera and merged it into the aerial AVM image. However, the
slow processing speed of the merging process is not suitable
for the real-time parking system.

To solve this problem, the proposed system employed
a look-up table (LUT) containing mapping relationships
between AVM images and fisheye-camera images. After
storing the ROI of the AVM image that needs to be calculated
in advance to the LUT, the system performed parking-line
detection by approaching the pixel value corresponding to
the coordinates of the fisheye image in the LUT. Realtime
performance is guaranteed by significantly reducing
processing time owing to the manner in which information is
stored in the LUT and read only when necessary.

5) Parking

Parking is implemented in five steps. The first step is to
locate the vacant parking space after positively identifying it
through the aforementioned process. In this step, the steering
angle of the wheels is changed in the opposite direction of
the parking bay. The vehicle is then driven forward a certain
distance—approximately 1.8 m depending on the type of
vehicle. To keep within 1.8 m, the vehicle is driven while
comparing the current location of the vehicle with the destination
indicated after mapping (the destination was mapped
1.8 m away in the SLAM map). The scale between SLAM
maps and topographical maps is required for proper mapping.
To address this, the system began computing the scale at the
start of the drive.

\begin{figure}[htp] 
    \centering 
    \includegraphics[width=240pt]{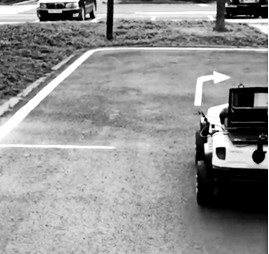} 
    \caption{Line for scale calculation} 
    \label{fig:fig14}
\end{figure}

One line was placed on the left side of the starting position
as shown in Fig. 14. This line on the left is called the scale
line. As the vehicle moves forward at a constant speed, the
scale line is detected by the parking-line detection algorithm.

\begin{figure}[htp] 
    \centering 
    \includegraphics[width=240pt]{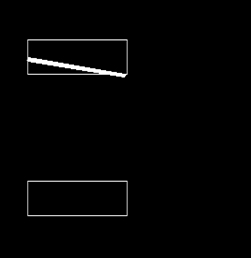} 
    \caption{Result of scale-line detection} 
    \label{fig:fig15}
\end{figure}

Scale lines are detected in aerial images as shown in Fig.
15. When the scale line crosses two pre-specified areas in the
aerial image, the SLAM map displays the coordinates of the
vehicle at these points. When the vehicle is past the scale
line, the SLAM map shows coordinates of two vehicles. The
scale ratio between the aerial image and the SLAM map is
calculated using the length difference between this pair of
coordinates and the distance between a pre-specified area in
the aerial image.

This information is saved and driving continues. If a parking
lot is detected while driving, the aerial image will show the
parking line as described earlier. Comparing the length of the
parking line shown in the aerial image with the length of the
actual parking line, the scale between the topographical and
aerial images is also calculated.

The scale between the SLAM map and the aerial image
and that between the aerial image and the topographical map 
ultimately allows for the derivation of the scale between the
SLAM map and the topographical map. This scale allows the
SLAM map to display the previously determined destination
approximately 1.8 m away. After the vehicle moves to a
destination that is a certain distance away, the third step of
parking is carried out. In the third step, the vehicle’s wheels
are turned in the direction of the parking bay and the vehicle
is then reversed. If the vehicle is parallel to the parking line
while reversing, the steering angle of the vehicle is restored
so that the vehicle’s wheels are flush with the body of the
vehicle (i.e., parallel to the parking line as well). The fisheye
cameras attached to the left and right of the vehicle are used for
determining whether the vehicle is parallel to the parking line.
The slope of the detected lines is obtained in advance when
the vehicle is parallel to the parking line. The slope of the
lines in the left/right camera image is then compared with the
slope obtained in advance while reversing. If the two slopes are
identical, the vehicle is considered to be parallel to the parking
line. If the vehicle reverses parallel to the parking line, it starts
to intelligently determine whether it has entered the parking
bay exactly; images obtained from the rear camera are used
for this operation. The rear camera continuously compares
the gap between the vehicle and the line behind the parking
bay (i.e., the base line) while continuing to detect the line.
When this gap decreases below a certain threshold, the vehicle
is correctly located inside the parking lot and the vehicle is
stopped. The parking algorithm is then terminated.

\section{EXPERIMENT}

\subsection{Implementation Environment} The environment used to implement the proposed system
comprised an Intel (R) Core (TM) i7-7700HQ CPU, 8.00 GB
RAM, and Ubuntu 16.04 OS. The experiment was conducted
both indoors and outdoors to simulate an actual parking
lot. Additionally, the road markings and parking bays were
reduced in proportion to the vehicle. Furthermore, the distance
used in the parking algorithm was also calculated and adjusted
to the rotation angle and radius of the HENES T870 model.
Finally, a nucleo board was used to send vehicle control
signals.

\subsection{Evaluation}

1) Autonomous Driving after Loop Closing

The control algorithm after loop closing presented in this
paper was evaluated by measuring the distance covered by
and steering-angle difference of the left front wheel when the
vehicle returned to its original position. The experiment was
performed five times at each of the three locations shown in
Fig. 16.

\begin{figure}[htp] 
    \centering 
    \includegraphics[width=100pt]{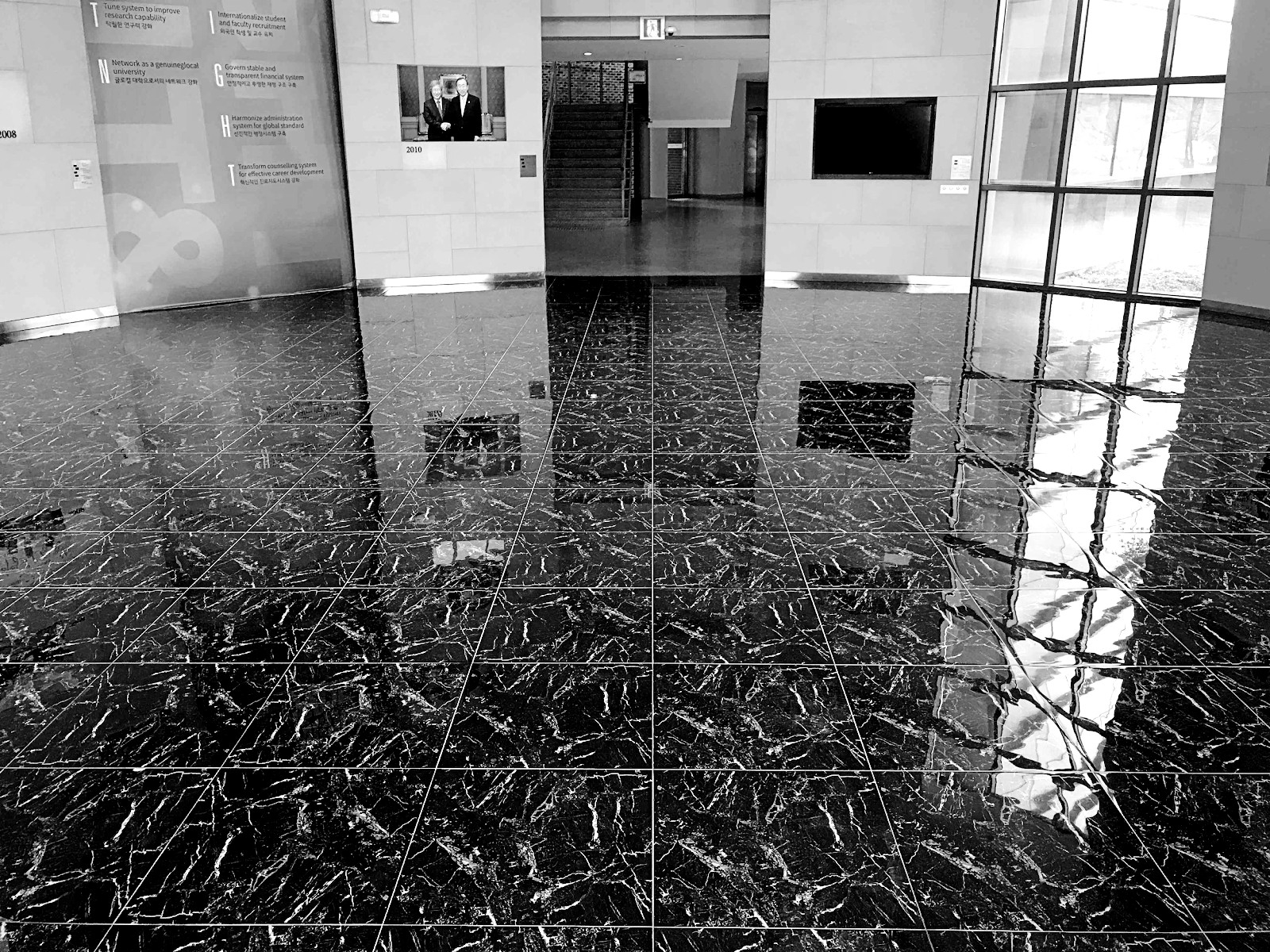} 
    \includegraphics[width=100pt]{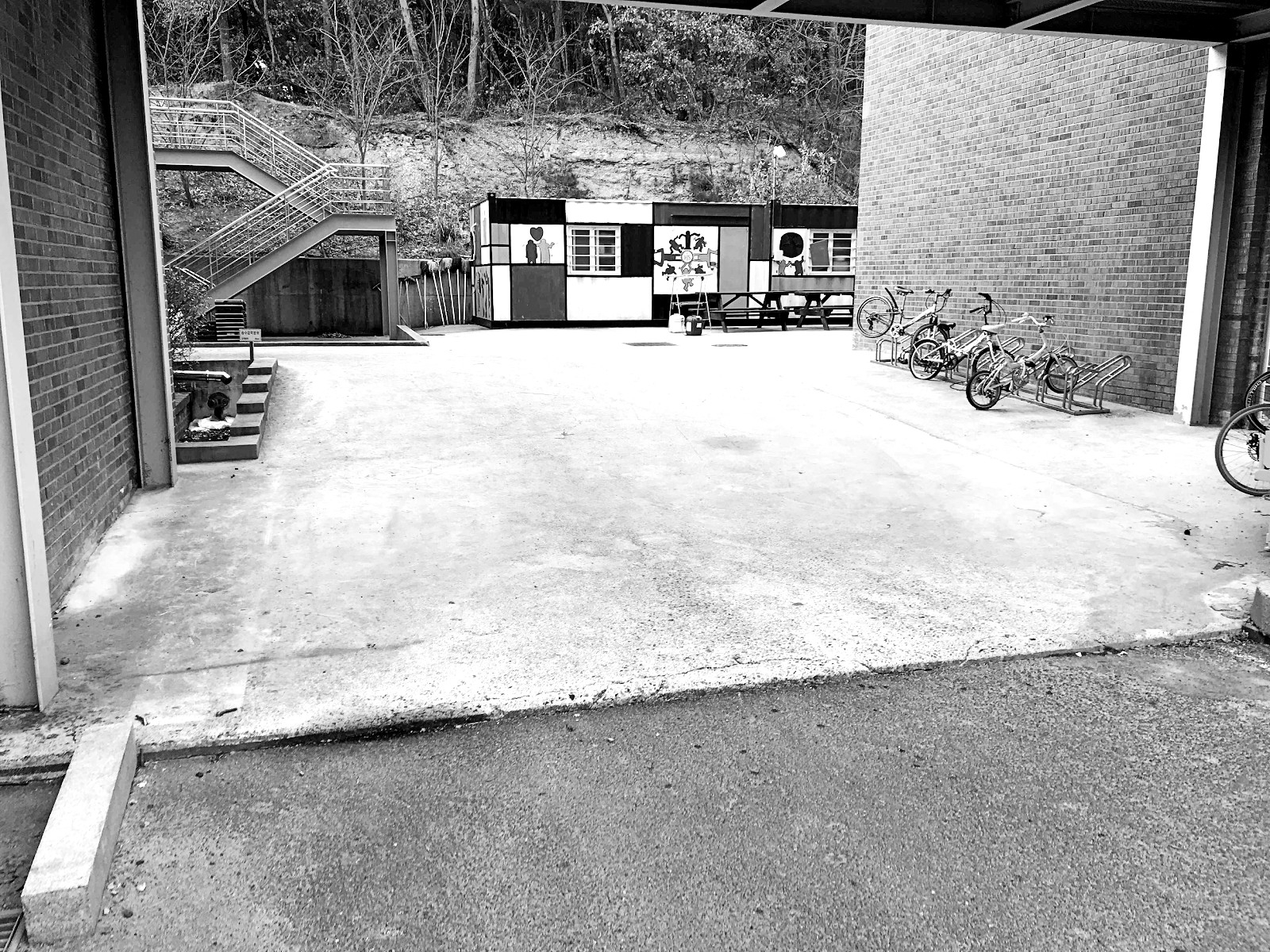} 
    \includegraphics[width=100pt]{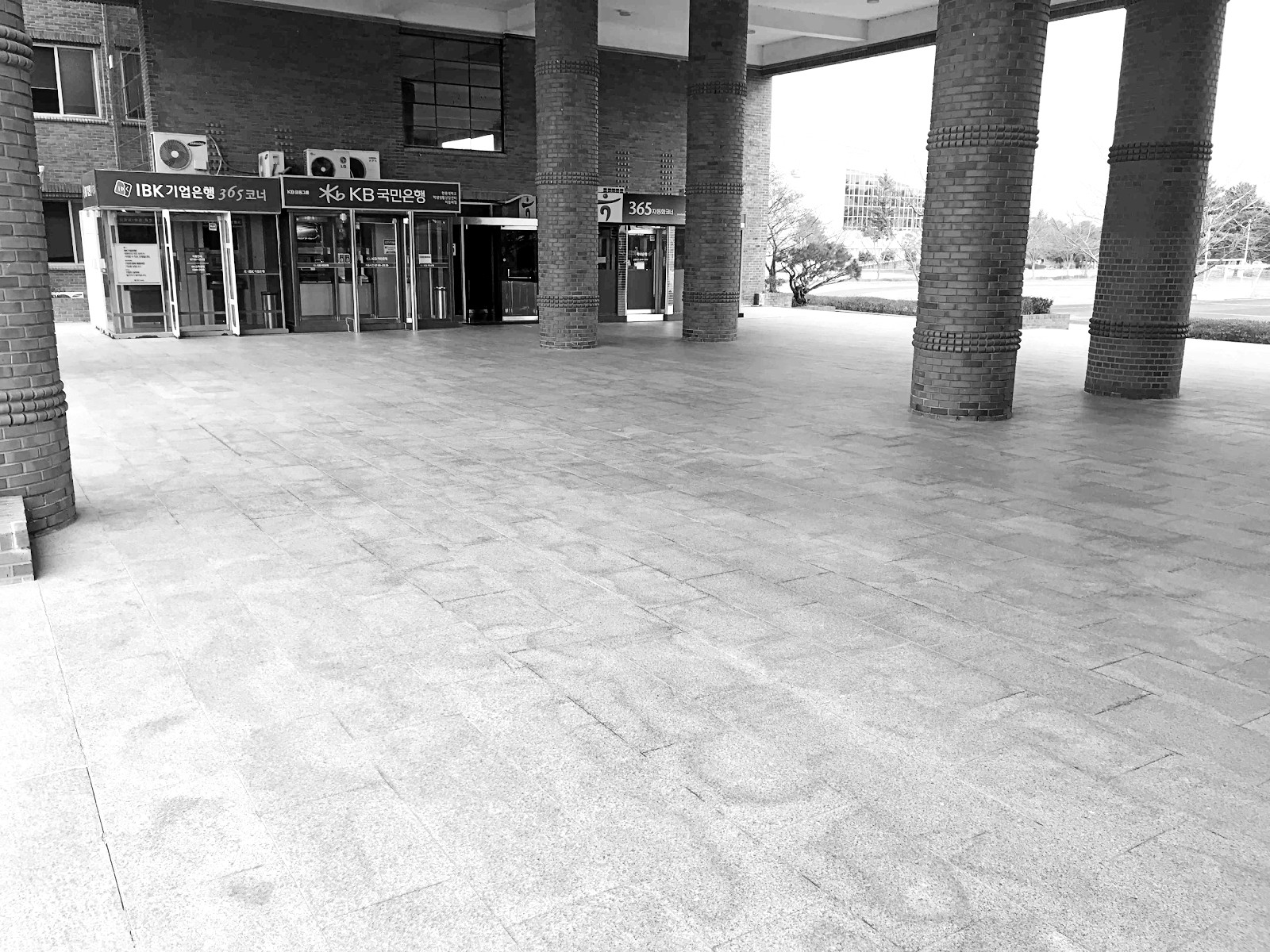} 
    \caption{(a) A narrow indoor space with light reflection on the floor; (b)
A narrow outdoor space; and (c) A large outdoor space.} 
    \label{fig:fig16}
\end{figure}

\begin{table}[htp]
\caption{Distance Difference Measurement}
\begin{center}
\begin{tabular}[width=8.5cm]{c c c c c c}
\hline\hline
    \multirow{2}{*}{\textbf{Place}} & \multicolumn{5}{c}{\textbf{Distance Difference (cm, \textpm0.5)}} \\ 
        & \textbf{Trial 1} & \textbf{Trial 2} & \textbf{Trial 3} & \textbf{Trial 4} & \textbf{Trial 5} \bigstrut \\
\hline
    \begin{tabular}[c]{@{}c@{}}a\end{tabular} & 8 & 5 & 11 & 8 & 5 \bigstrut \\
    \begin{tabular}[c]{@{}c@{}}b\end{tabular} & 8 & 5 & 6 & 6 & 8 \bigstrut \\
    \begin{tabular}[c]{@{}c@{}}c\end{tabular} & 5 & 7 & 12 & 8 & 11 \bigstrut \\
\hline\hline
\end{tabular}
\label{table1}
\end{center}
\end{table}

\begin{table}[H]
\caption{Angle Difference Measurement}
\begin{center}
\begin{tabular}[width=8.5cm]{c c c c c c}
\hline\hline
    \multirow{2}{*}{\textbf{Place}} & \multicolumn{5}{c}{\textbf{Angle Difference (\textdegree, \textpm2.5)}} \\ 
        & \textbf{Trial 1} & \textbf{Trial 2} & \textbf{Trial 3} & \textbf{Trial 4} & \textbf{Trial 5} \bigstrut \\
\hline
    \begin{tabular}[c]{@{}c@{}}a\end{tabular} & 10 & 15 & 20 & 15 & 15 \bigstrut \\
    \begin{tabular}[c]{@{}c@{}}b\end{tabular} & 15 & 15 & 10 & 15 & 20 \bigstrut \\
    \begin{tabular}[c]{@{}c@{}}c\end{tabular} & 15 & 10 & 15 & 15 & 10 \bigstrut \\
\hline\hline
\end{tabular}
\label{table2}
\end{center}
\end{table}

The results in (a) and (b) are very similar, whereas the
results in (c) are slightly pronounced. This is attributed to
the environment of (c) being larger than the other two environments.
Furthermore, the average distance that had to be
covered driving in a straight line was greater in (c) compared
to that in (a) and (b). Keyframes were generated when the
level of new features exceeded that exhibited in previous
keyframes. However, similar features were continuously detected
in straight routes. Therefore, there was a large distance
gap between keyframes in straight routes. Interestingly, for
straight routes, the distance between adjacent keyframes did
not affect the autonomous driving. Nevertheless, the error in
driving was negligible even in the case of (c).

The above results show that the algorithms presented by
this study exhibit high accuracy for both indoor and outdoor
environments. In other words, the error is not significant
when the vehicle drives along the same path it did before the
loop closing occurred, without capturing any new information
but instead utilizing the information in the SLAM map.

2) Arrow Recognition Rate Experiment
The experiment was conducted by reducing the size of the
actual road markings by the same reduction ratio of the parking
lot space. It was conducted on three regular outdoor pavements
and one indoor area, with a total of 20 experimental processes,
each mutually exclusive for a left or right arrow.

\begin{table}[H]
\caption{Success Rate of Arrow Recognition}
\begin{center}
\begin{tabular}[width=8.5cm]{c c c c}
\hline\hline
    \multicolumn{4}{c}{\textbf{Place}} \\ 
    \textbf{Outdoor 1} & \textbf{Outdoor 2} & \textbf{Outdoor 3} & \textbf{Indoor} \bigstrut \\
\hline
     85\% & 80\% & 90\% & 70\% \bigstrut \\
\hline\hline
\end{tabular}
\label{table3}
\end{center}
\end{table}

The indoor environment presented a lower success rate
than the outdoor environments. This may be because the glare
on the floor caused by indoor lighting was detected during
the edge detection process, hindering the correct detection
of arrows. If a robust method of illumination interference is
applied in the system, it may present higher success rates in
the indoor environment.

3) Parking Success Rate Experiment

In this experiment, we evaluated the success of the parking
algorithm based on whether the vehicle entered the parking
bay correctly after locating the bay. The experiment was
conducted indoors with glare on the floor and outdoors on
asphalt to measure the accuracy of parking-line recognition
depending on the environment. The parking bay was located
only on the left side of the vehicle for repeated experiments
under the same conditions.

\begin{table}[H]
\caption{Success Rate of Parking Algorithm}
\begin{center}
\begin{tabular}[width=8.5cm]{c c}
\hline\hline
    \multicolumn{2}{c}{\textbf{Place}} \\ 
    \textbf{Outdoor} & \textbf{Indoor} \bigstrut \\
\hline
     75\% & 62.5\% \bigstrut \\
\hline\hline
\end{tabular}
\label{table4}
\end{center}
\end{table}

After executing the algorithm in each case, it was considered
as a success if the distance between the nearest border and
the vehicle exceeded 18 cm; otherwise, it was considered as a
failure. This criterion was established based on the condition
when the vehicle was located in the center of the parking area
parallelly, considering the size of the parking bay.

The experiments showed 62.5% accuracy indoors and
75% accuracy outdoors. Parking-line recognition and template
matching were performed effectively in both environments.
Parallel alignment and detection of the end of the parking algorithm
were also successfully performed in both environments.
However, when the vehicle could not move to the appropriate
location to reverse, the autonomous parking was unsuccessful
as shown in Fig. 17. This may have resulted from poor scale
estimation from SLAM and topographical maps.

\begin{figure}[htp] 
    \centering 
    \includegraphics[width=90pt]{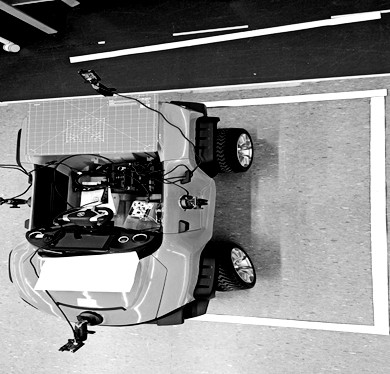} 
    \includegraphics[width=90pt]{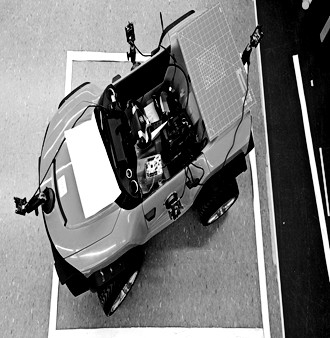} 
    \includegraphics[width=90pt]{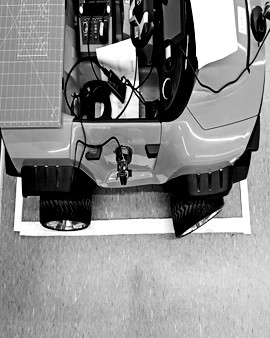} 
    \caption{Failure cases in indoor environment} 
    \label{fig:fig17}
\end{figure}

\section{CONCLUSION}

In this paper, we proposed a system that implements autonomous
parking using only camera sensors. The proposed
self-driving method involves detecting parking bays and direction
arrows via a fisheye camera at the front, rear, left, and
right of the vehicle. During self-driving, the SLAM map was
constructed by identifying the features in the image of the
front camera. AVM images were utilized for parking¬ space
detection. In this approach, we reduced the computational cost
by using the LUT of the two pre-acquired images, rather
than by directly converting the fisheye image into the AVM
image. Then, when a parking space was detected, parking
was executed by different parking algorithms depending on
the location of the vehicle with respect to the parking space
shown on the SLAM map. If the left, right, and rear cameras
detected that the vehicle had entered the parking line correctly,
parking maneuvers were halted.

Considering that driving route data are stored using SLAM
technology, a system can be developed in the future to return
the vehicle to the location where parking was initiated.
Notwithstanding, the calculation of scale has not been completely
automated. Moreover, the success rate of autonomous
parking is not at a level sufficient for comer-cialization. This
seems to be a limitation arising from the heavy reliance on
cameras. In this regard, if we apply sensor fusion using other
sensors such as ultrasound sensors in the system, we can
expect higher accuracy and automation compared with that
of the system proposed in this paper.

In conclusion, this paper contributes to the field of autonomous
parking on account of asserting that it is possible
to develop a self-parking system to a level where completely
autonomous parking is possible.

%%%%%%%%%%%%%%%%%%%%%%%%%%%%%%%%%%%%%%%%%%%%%%%%%%%%%%%%%%%%%%%%%%%%%%%%%%%%%%%%

References are important to the reader; therefore, each citation must be complete and correct. If at all possible, references should be commonly available publications.

\begin{biography}
[{\includegraphics[width=1in,height=1.25in,clip,keepaspectratio]{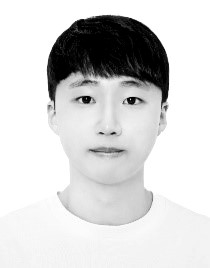}}]
{Young Gon Jo} graduated from Handong University in 2021 with the B.S. degree of computer science and electric engineering.

He currently works for the algorithm team at VADAS in Pohang, Republic of Korea. His research interest includes the image processing and autonomous driving technology.
\end{biography}

\begin{biography}
[{\includegraphics[width=1in,height=1.25in,clip,keepaspectratio]{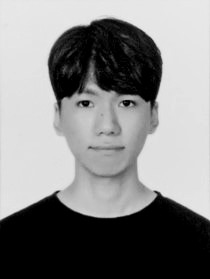}}]
{Seok Hyeon Hong} is currently pursuing the B.S. degree in computer science and engin- eering with the School of Computer Science and Electrical Engineering, Han-dong University.

Since 2019, he has been a Research Assistant with the Computer Graphics and Vision Laboratory. His current research interests include computer vision and computer graphics.
\end{biography}

\begin{biography}
[{\includegraphics[width=1in,height=1.25in,clip,keepaspectratio]{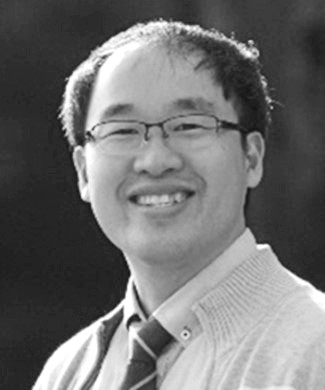}}]
{Sung Soo Hwang} graduated from Handong University in 2008 with a B.S. degree of electric engineering and com-puter science. He received the M.S. and Ph.D. degree in electrical and electronic engineering from the Korea Advanced Institute of Science and Technology in 2010 and 2015, respectively.

He is currently an associate professor at Handong University in the School of Computer Science and Electrical Engineering. His research interest includes the creation and operation of video maps for augmented reality and autonomous driving.
\end{biography}

\begin{biography}
[{\includegraphics[width=1in,height=1.25in,clip,keepaspectratio]{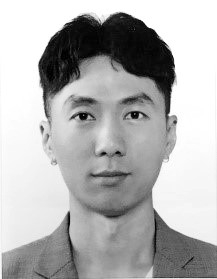}}]
{Jeong Mok Ha} received his B.S. degree in electrical engineering from the Pusan National University, Busan, Republic of Korea, in 2010 and Ph.D. in electrical engineering from the Pohang University of Science and Technology (POSTECH), Pohang, Republic of Korea, in 2017.

He is currently an algorithm team leader in VADAS Co., Ltd., Pohang, Republic of Korea. He is interested in automotive vision, including camera calibration, surround view, deep learning, and SLAM.
\end{biography}

\end{document}